\documentclass[conference]{IEEEtran}
\IEEEoverridecommandlockouts
\usepackage{amsmath,amssymb,amsfonts}
\usepackage{algorithmic}
\usepackage{graphicx}
\usepackage{textcomp}
\usepackage{xcolor}
\usepackage[numbers]{natbib} 
\usepackage{placeins}
\usepackage{multirow}
\usepackage{multicol}
\def\BibTeX{{\rm B\kern-.05em{\sc i\kern-.025em b}\kern-.08em
		T\kern-.1667em\lower.7ex\hbox{E}\kern-.125emX}}
\begin{document}
	
	\title{Zero-shot Speech Translation \\
		\thanks{This thesis was prepared in partial fulfilment of the requirements for the Degree of Bachelor of Science in Data Science and Artificial Intelligence, Maastricht University. Supervisor(s): Dr. Jan Niehues}
	}
	
	\author{\IEEEauthorblockN{Tu Anh Dinh}
		\IEEEauthorblockA{\textit{Department of Data Science and Knowledge Engineering} \\
			\textit{Maastricht University}\\
			Maastricht, The Netherlands}
	}
	
	\maketitle
	
	\begin{abstract}
		Speech Translation (ST) is the task of translating speech in one language into text in another language. Traditional cascaded approaches for ST, using Automatic Speech Recognition (ASR) and Machine Translation (MT) systems, are prone to error propagation. End-to-end approaches use only one system to avoid propagating error, yet are difficult to employ due to data scarcity. We explore zero-shot translation, which enables translating a pair of languages that is unseen during training, thus avoid the use of end-to-end ST data. Zero-shot translation has been shown to work for multilingual machine translation, yet has not been studied for speech translation. We attempt to build zero-shot ST models that are trained only on ASR and MT tasks but can do ST task during inference. The challenge is that the representation of text and audio is significantly different, thus the models learn ASR and MT tasks in different ways, making it non-trivial to perform zero-shot. These models tend to output the wrong language when performing zero-shot ST. We tackle the issues by including additional training data and an auxiliary loss function that minimizes the text-audio difference. Our experiment results and analysis show that the methods are promising for zero-shot ST. Moreover, our methods are particularly useful in the few-shot settings where a limited amount of ST data is available, with improvements of up to +11.8 BLEU points compared to direct end-to-end ST models and +3.9 BLEU points compared to ST models fine-tuned from pre-trained ASR model.
	\end{abstract}
	
	
	\section{Introduction} \label{sec:intro}
	Speech Translation (ST) is the task of translating speech audio in a source language into text in a target language. Traditional approaches for ST include two cascaded steps: Automatic Speech Recognition (ASR) and Machine Translation (MT). ASR transcribes speech into text of the same language, and MT translates the text output by ASR to the text in the target language. These approaches are prone to errors propagated from the ASR step to the MT step \cite{CascadedError}. Due to that, end-to-end Speech Translation has recently been gaining more interest. In end-to-end ST, the speech audio in a source language is translated directly into text in a target language and error propagation is no longer an issue.  The challenge with end-to-end ST is the lack of appropriate end-to-end data, i.e, samples of speech in the source language and the corresponding text translation in the target language \cite{sperber-paulik-2020-speech}.
	
	To tackle the data scarcity issue, Zero-shot Speech Translation will be explored in this paper. Zero-shot, in the context of translation models, is an approach that enables translating a pair of languages even when no end-to-end data of that particular pair was observed during training \cite{sperber-paulik-2020-speech}. Zero-shot has been shown to work for multilingual machine translation, i.e., translating unseen pairs of languages in text format \cite{googleNMT}. We attempt to build zero-shot models for speech translation: models that are trained only on ASR and MT tasks but can do ST task during inference. Here only one model is used to perform ST task during inference, which makes zero-shot ST an end-to-end approach. As an example, a model is trained with samples of English audio to English text (an ASR task) and samples of English text to German text (an MT task). Using zero-shot approaches, the model is expected to have the ability to translate English audio to German text (an ST task), which has not been seen in the training data. Thus, zero-shot ST avoids the use of end-to-end ST data for training. 
	
	We encounter several challenges when building zero-shot ST models. First, the zero-shot models using only ASR and MT training data output the wrong language while performing ST task (they output the source language instead of the target language). Second, the difference in representation of text and audio makes the models learn the ASR and MT tasks in different ways, which makes it non-trivial to perform zero-shot. We study and propose several approaches to tackle the issues: (1) removing the residual connections of a middle encoder layer to encourage language-independent representation of the data \cite{DEPI}; (2) using an auxiliary loss function to minimize the difference in text and audio representation \cite{ZS-Quan}; (3) using augmented training data to force the models to learn to output the correct language and (4) using additional opposite training data also to force outputting the correct language. We find that the first approach, which was originally used for zero-shot multilingual MT, does not work for zero-shot ST. The other approaches prove to be promising, although they have not improved zero-shot ST to the point that it can be used practically. We also find these approaches to be particularly useful in the few-shot setting, when a limited amount of ST data is available, by comparing our few-shot models to some end-to-end baselines using a similar model architecture. Our few-shot models outperform direct end-to-end ST models, i.e., models trained on the same amount of ST data from scratch, by up to +11.8 BLEU points. Our models also outperform ST models fine-tuned from a pre-trained ASR model using the same amount of ST data by up to +3.9 BLEU points. 
	
	Our study addresses the following research questions. (1) How data-efficient are end-to-end and cascaded models? (2) Can techniques from zero-shot multilingual MT be applied to end-to-end ST? (3) How can we model the different modalities in zero-shot ST?
	
	The rest of this paper is structured as follows. Section \ref{sec:RelatedWork} discusses the related work. Section \ref{sec:zsST} describes our approaches for zero-shot ST. Section \ref{sec:exp_setup} explains our experiment setup; Section \ref{sec:results} discusses the experiment results. We further provide an in-depth analysis of our models in Section \ref{sec:analysis}. Finally, Section \ref{sec:conclusions} draws conclusions and outlines future work.

	\section{Related work} \label{sec:RelatedWork}
	In this section, we discuss existing work on Speech Translation (ST) as well as Zero-shot translation.
	
	Different approaches for ST have been explored over the decades, as summarized in \cite{sperber-paulik-2020-speech}. We highlight some related points in \cite{sperber-paulik-2020-speech} as follows. Cascaded ST is the traditional approach, which was first introduced in \cite{IntroCascade}. Cascaded ST uses two separately-built systems: ASR and MT. Cascaded ST is a strong approach thanks to the well-established research and the abundance of data in ASR and MT. One shortcoming of cascaded ST is error propagation \cite{CascadedError}. Therefore, more attempts have been made towards the end-to-end approach, which uses only one system to perform ST. Examples of end-to-end ST include direct ST models trained on end-to-end ST data from scratch \cite{ExampleDirect}, models pre-trained on an ASR task and fine-tuned on ST task \cite{bansal-etal-2019-pre}, models trained on ST data generated by augmenting ASR or MT data \cite{jia2019leveraging, pino2019harnessing} and models co-trained on MT and ST task \cite{tang2021general}. However, a major challenge of end-to-end ST is the lack of appropriate data. The available ST data are very limited in size and language coverage. For this reason, despite the efforts, many end-to-end ST approaches have not been able to outperform the traditional cascaded approach.
	
	
	For zero-shot translation, it has been shown to work for multilingual  MT. In \cite{googleNMT}, zero-shot multilingual MT is enabled by adding a language token to the beginning of the input sequence to indicate the required target language. The authors in \cite{ZSha2016} enable zero-shot multilingual MT by additionally concatenating language codes to every word. Several studies have been done to improve the quality of zero-shot multilingual MT. In \cite{ZSha2017}, target language embeddings are used as a model feature instead of modifying the raw data to reduce vocabulary size. Multiple approaches to encourage a source-language-independent representation are proposed in \cite{ZS-Quan}, such as using a fixed size encoder for different languages. In \cite{DEPI}, a language-independent representation is encouraged by disentangling positional information of the input and output tokens.
	
	Inspired by the ability of zero-shot learning for multilingual MT, we study the applicability of similar approaches on Speech Translation, described in the following section. 
	
	\section{Zero-shot Speech Translation} \label{sec:zsST}
\begin{figure}[htbp]
		\centerline{\includegraphics[width=0.4\textwidth]{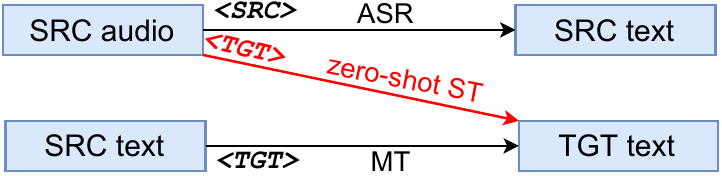}}
		\caption{Idea for zero-shot ST. Black arrows are training directions. Red arrow is zero-shot inference. Tags in the brackets are target-language tokens.}
		\label{fig:baseZS}
	\end{figure}

	Fig. \ref{fig:baseZS} illustrates the proposed approach for training a zero-shot ST model. The model is trained on two tasks simultaneously: Automatic Speech Recognition (ASR) and Machine Translation (MT). ASR training data include samples of audio in a source language and the corresponding text in the same language (\textit{SRC audio} $\rightarrow$ \textit{SRC text}). MT training data include samples of text in the source language and the corresponding text in a target language (\textit{SRC text} $\rightarrow$ \textit{TGT text}). Using zero-shot, the model is expected to be able to perform  ST task during inference, i.e., translating \textit{SRC audio} $\rightarrow$ \textit{TGT text}. In order for zero-shot to work, it is necessary that the model represents \textit{SRC audio} and \textit{SRC text} in a similar way so that it can leverage the ASR and MT tasks learnt during training to perform the ST task during inference.
	
\begin{figure}[htbp]
		\centerline{\includegraphics[width=0.52\textwidth]{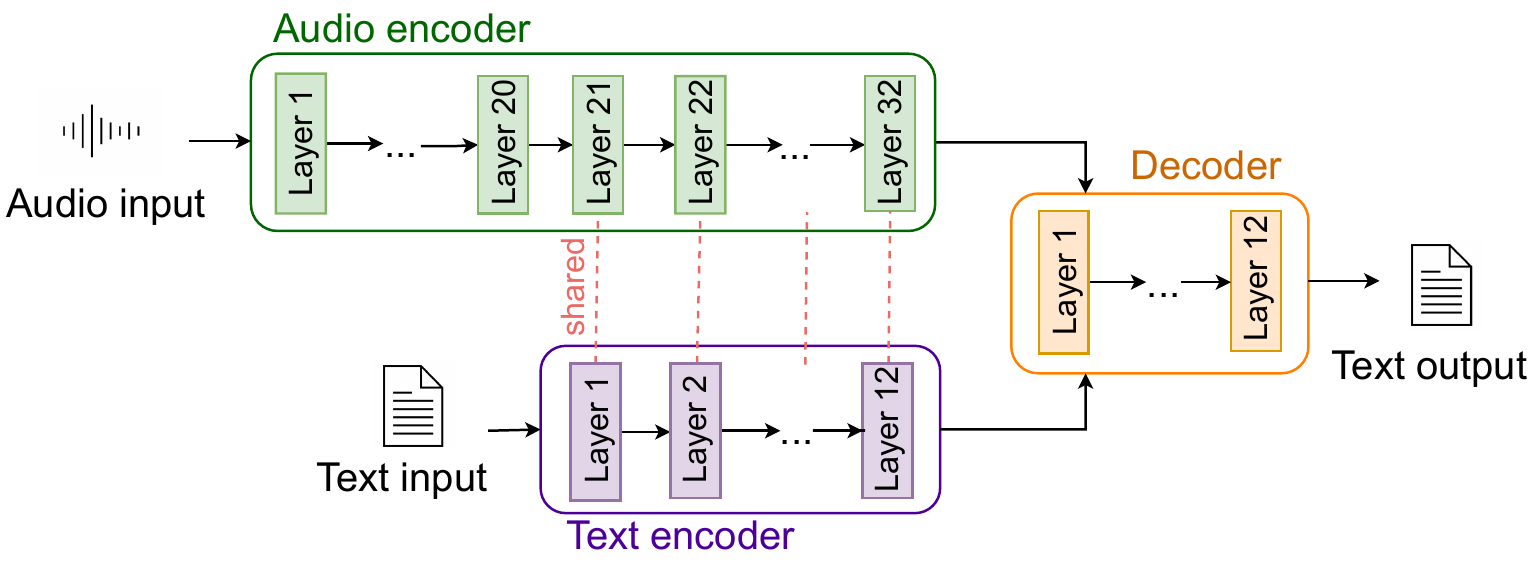}}
		\caption{Overall structure of multi-modality models.}
		\label{fig:OverallStructure}
	\end{figure}

	We use the Transformer architecture as described in \cite{attention} and \cite{quan-2019-very-deep}, with the attention-based encoder and decoder. We extend it by including two parallel encoders, one for text input and one for audio input to fit our multi-modality training data. All layers of the text encoders are shared with the last layers of the audio encoder, similar to the ones in \cite{tang2021general}. The overall structure is shown in Fig. \ref{fig:OverallStructure}. To enable zero-shot, we apply the same method as \cite{ZS-Quan}, which was originally used for zero-shot multilingual MT. We add target-language tokens to the beginning of input sequences and concatenate the target language embeddings to every decoder input to enforce the model outputting the language of interest. As can be seen in Fig. \ref{fig:baseZS}, with the same \textit{SRC audio} input, if we want the model to output \textit{SRC text}, we add the target-language token \textit{$<$SRC$>$} to the input; if we want the model to output \textit{TGT text}, we add the target-language token \textit{$<$TGT$>$} to the input. 
	
	We observe that this plain zero-shot model ignores the target-language tokens during training and is unable to do zero-shot ST (which is discussed further in section \ref{sec:baseZS}). Hence, the rest of this section describes the approaches to further encourage the zero-shot ability of the model.
	
	\subsection{Disentangling Positional Information} \label{sec:DEPI}
	The objective of a zero-shot ST model is to be able to switch between outputting \textit{SRC} and \textit{TGT} language based on the specified target-language token. Because of that, it is useful for the model to learn language-independent representation of the data. Therefore, we apply the  Disentangling Positional Information approach introduced in \cite{DEPI}. This approach was originally used to improve zero-shot multilingual MT. The idea is to relax the strong positional correspondence of the output to input tokens, hence give the model more freedom on word reordering to encourage a language-independent representation of the data. The authors in \cite{DEPI} achieved this by removing residual connections in a middle encoder layer. 
	
	As described earlier, our model has two parallel encoders, one for text and one for audio input. All layers of the text encoders are shared with the last layers of the audio encoder. Using the Disentangling Positional Information approach, we remove the residual connections in the middle layer of the text encoder, which is also shared with the audio encoder.
	
	\subsection{Auxiliary loss function} \label{sec:auxloss}
\begin{figure}[htbp]
		\centerline{\includegraphics[width=0.4\textwidth]{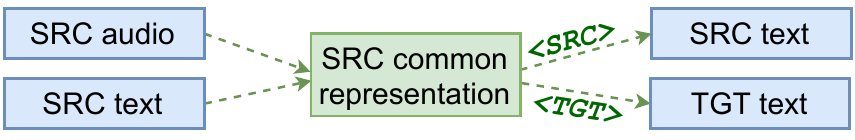}}
		\caption{Motivation for auxiliary loss. By minimizing text-audio difference, the model is expected to learn a modality-independent representation of the data (the green box). The tags in angle brackets are the target-language tokens.}
		\label{fig:auxloss}
	\end{figure}

	Another way to encourage zero-shot is to use an auxiliary loss function that minimizes the difference between encoder output of audio and text sentences. Using this auxiliary loss function, the model is expected to learn a modality-independent representation of the data, as shown in Fig. \ref{fig:auxloss}. If we have a sample sentence in text form and that same sentence in spoken (audio) form, the model should be able to represent the two samples similarly. In this way, it is expected that the model will be able to switch between outputting different languages during inference, hence encourage zero-shot. The metric chosen for the encoder output difference is the squared error of mean-pool over time, similar to the one suggested in \cite{ZS-Quan}. That is, given a pair of aligned text and audio sentences, for each sentence's encoder output, we take the average over time of the encoder state (i.e., mean-pooling), then calculate the squared error between the two mean-pooled vectors.
	
	\subsection{Data augmentation} \label{sec:data_aug}
	As described at the beginning of this section, we use target-language tokens to enforce the model outputting the language of interest. To be able to perform zero-shot, the model must learn the target-language tokens during training. It is observed that the plain model shown in Fig. \ref{fig:baseZS} ignores the language tokens during training and unable to do zero-shot ST. For \textit{SRC audio} input, even if we set the target token to be \textit{$<$TGT$>$}, the model still outputs \textit{SRC text}. One hypothesized reason is as follows. When training with only ASR and MT tasks as in Fig. \ref{fig:baseZS}, audio input always has \textit{SRC} language output and text input always has \textit{TGT} language output. Since the representation of text and audio input is different, the model ignores the target-language tokens and decides on the output language based on the modality instead, i.e., outputting \textit{SRC text} for all audio input and outputting \textit{TGT text} for all text input. 
	
	To tackle this issue, we create artificial data to be trained along with the main ASR and MT data, avoiding the need of searching for another real dataset with a real language. The amount of artificial data used is about half of the main training data. The idea is that input in each modality (text and audio) should have more than one target language output during training. In that way, the model will be forced to look at the target-language tokens to decide which language to output. 
	
\begin{figure}[htbp]
		\centerline{\includegraphics[width=0.4\textwidth]{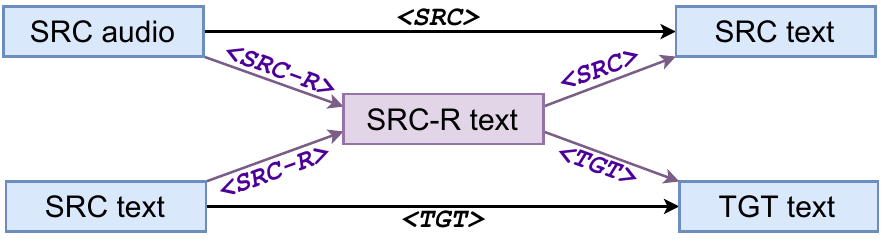}}
		\caption{Augmented training data. The black arrows are the main training directions. The purple arrows are the artificial directions.}
		\label{fig:AD4}
	\end{figure}

	We create artificial training data by augmenting the ASR and MT data that we already have.  We introduce an artificial language created from the \textit{SRC} language as follows:
	\begin{itemize}
		\item Reverse source language sentences character-wise
		\item Lowercase the letters and strip away all punctuations
		\item Restore the common rules: capitalize the beginning of sentences, punctuation at the end of sentences
	\end{itemize}
	For example, "Hello world!" in English will be “Dlrow olleh!” in reversed English. We denote the artificial language as \textit{SRC-R}. The set of training data is shown in Fig. \ref{fig:AD4}. In addition to the main ASR and MT samples, we add \textit{SRC audio} $\rightarrow$ \textit{SRC-R text} and \textit{SRC text} $\rightarrow$ \textit{SRC-R text} samples. In this way, \textit{SRC audio} and \textit{SRC text} now have two target languages, which forces the model to learn the language tokens to know which language to output. We also include \textit{SRC-R text} $\rightarrow$ \textit{SRC text} and \textit{SRC-R text} $\rightarrow$ \textit{TGT text} samples so that the model can learn to switch between outputting \textit{SRC} and \textit{TGT} languages. 
	
	\subsection{Additional opposite training data}
	Recall that our main training data are ASR and MT data of \textit{SRC} $\rightarrow$ \textit{TGT} direction. In addition to the main training data, we include ASR and MT data of the opposite direction \textit{TGT} $\rightarrow$ \textit{SRC}. The motivation here is the same as in Section \ref{sec:data_aug}: input in each modality should have more than one target language output during training. Hence, the training data are:
	\setlength\multicolsep{3pt}
	\begin{multicols}{2}
		\begin{itemize}
			\item \textit{SRC audio} $\rightarrow$ \textit{SRC text}
			\item \textit{SRC text} $\rightarrow$ \textit{TGT text}
			\item \textit{TGT audio} $\rightarrow$ \textit{TGT text}
			\item \textit{TGT text} $\rightarrow$ \textit{SRC text}
		\end{itemize}
	\end{multicols}
	Here, each input modality (text and audio) has two target languages: \textit{SRC} and \textit{TGT}, which encourages the model to learn the target-language tokens. One disadvantage of this approach is that it requires the audio data of the \textit{TGT} language.
	
	\subsection{Few-shot Speech Translation}
	As mentioned previously, the issue with end-to-end ST models is the lack of ST data for training. In addition to building zero-shot ST models, which require no ST data for training, we build few-shot models by fine-tuning the zero-shot models with a small amount of ST data to see how the models perform when a limited amount of ST data is available. It is expected that the fine-tuning process with ST data will help boosting the ST performance compared to zero-shot models.
	
	\section{Experimental setup} \label{sec:exp_setup}
	The goal of the experiments is to build and evaluate zero-shot ST models to translate English audio to German text, denoted as \textit{EN audio} $\rightarrow$ \textit{DE text}.
	
	\subsection{Dataset and preprocessing}
	We use the CoVoST 2 dataset \cite{wang2020covost}, a multilingual speech-to-text translation corpus including speech audio samples and the corresponding transcription and translation of different language pairs. The audio contains short spoken sentences with an average length of 5 seconds. The transcriptions and translations have an average length of 9 words per sentence.
	
	The main data used in the experiments are English audio along with English transcription and German translation. The source language (\textit{SRC}) is \textit{EN}. The target language (\textit{TGT}) is \textit{DE}. This dataset contains 289K samples for training, 15K samples for validation and 15K samples for testing.
	We experiment with different portions of training data (10\%, 25\%, 33\% and 100\%). The validation set and test set are the same across all models, regardless of the amount of data used for training.
	Additionally, in some experiments, we also use German audio along with German transcription and English translation data, i.e, the \textit{DE $\rightarrow$ EN} dataset. The amount of the \textit{DE $\rightarrow$ EN} data used is always half of the amount of the \textit{EN $\rightarrow$ DE} data.

	%
	
	For audio data, we extract and normalize 40-dimensional log scale mel filterbank concatenated with its delta coefficient to use as input features. For text data, we remove the double quotes at the beginning and the end of the sentences if any, and use SentencePiece \cite{sentencepiece} without pre-tokenization and pre-normalization to build subword-based vocabularies. 
	
	\subsection{Model configurations}
	Our models use the Transformer architecture with attention-based encoder and decoder \cite{attention, quan-2019-very-deep}. For single-task models, we closely follow the hyperparameter choices in \cite{pham2019iwslt}. Models trained on a single audio-related task (ASR or ST) have 32 encoder layers, 12 decoder layers and death rate of 0.5. Models trained on a single MT task have 8 encoder layers, 8 decoder layers and no death rate. We adapt the hyperparameters to our mix-modality, multi-task models: 32 audio encoder layers, 12 text encoder layers, 12 decoder layers and no death rate; the 12 text encoder layers are shared with the last 12 audio encoder layers; shared weights for encoder and decoder word embeddings. All models use embedding size of 512, inner size of 2048, dropout rate of 0.2, attention dropout rate of 0.2, word dropout rate of 0.1, embedding dropout rate of 0.1, label smoothing rate of 0.1, Adam optimizer with learning rate of 0.01 and 8000 warm-up steps. 
	
	When training multiple datasets in one model (e.g., ASR and MT datasets), the batches from each dataset are ordered alternatively with a ratio such that each dataset is iterated through once in every epoch. Our models are trained for 64 epochs. One exception is the fine-tuned models, where we check the validation loss after every epoch and stop training as soon as the validation loss stops reducing. The checkpointed model with the lowest validation loss is used for final evaluation on the test set.
	
	\subsection{Evaluation metrics}
	For ASR tasks, the metric used is the Word Error Rate (WER) in percentage, calculated using VizSeq \cite{wang2019vizseq}. Before evaluation, the models' output transcriptions and the human-labeled transcriptions are lowercased, tokenized and the punctuation marks are removed, except for apostrophes and hyphens. The lower the WER, the better the performance.
	
	For MT and ST tasks, the metric used is the BLEU score, calculated using sacreBLEU \cite{bleu}. Before evaluation, the models' output is detokenized and the case is kept as it is to calculate case-sensitive BLEU score. The higher the BLEU score, the better the performance.
	
	\section{Results and Discussion} \label{sec:results}
	\subsection{Cascaded approach versus end-to-end approach}
	Table \ref{tab:CascadedVsEndToEnd} shows the results of training individual tasks with different amounts of the data as well as the performance of cascaded ST using ASR and MT models. With smaller amounts of data, ASR and MT tasks are easier to learn compared to ST task. With 25\% of the training data, we achieve a reasonable score for ASR and MT (43.6\% WER and 23.1 BLEU points, respectively), which is not the case for direct end-to-end ST (with 0.8 BLEU points). As a result, cascaded ST is much more data-efficient than direct end-to-end ST. Using 33\% of the training data, we obtain 14.0 BLEU points by cascaded ST, which is only 0.9 points less than using all of the data to train a direct end-to-end ST model. 
	
	\begin{table}[htbp]
		\caption{Performance of models trained on single tasks.}
		\begin{center}
			\begin{tabular}{|r|r|r|r|r|}
				\hline
				\multicolumn{1}{|c|}{\begin{tabular}[c]{@{}c@{}}Data portion\end{tabular}} &
				\multicolumn{1}{c|}{ASR} &
				\multicolumn{1}{c|}{MT} &
				\multicolumn{1}{c|}{Cascaded ST} &
				\multicolumn{1}{c|}{Direct end-to-end ST} \\ \hline
				25\%  & 43.6 & 23.1 & 11.5 & 0.8  \\ \hline
				33\%  & 37.5 & 26.3 & 14.0 & 1.6  \\ \hline
				100\% & 25.8 & 33.0 & 20.6 & 14.9 \\ \hline
			\end{tabular}
			\label{tab:CascadedVsEndToEnd}
		\end{center}
	\end{table}
	
	By this experiment, we confirm that zero-shot ST is an attractive approach. In zero-shot ST, we attempt to build a model that is trained on ASR and MT, which are the two data-efficient tasks, to perform ST during inference. Here only one model is trained and used to perform ST during inference, which makes zero-shot ST an end-to-end approach and avoids the error propagation issue of the cascaded approach. Since ASR and MT tasks have reasonable performance with only 25\% of the training data, we use 25\% of the data for the subsequent experiments and only train the most promising models on the full data.
	
	\subsection{Plain zero-shot Speech Translation models} \label{sec:baseZS}
	The performance of plain zero-shot models is shown in Table \ref{tab:baseZS}. We also include the performance of single-task models for comparison. Zero-shot models' MT performance is close to models trained on only MT task, with differences less than 0.4 BLEU  points. For ASR, zero-shot models' performance is slightly worse than models trained only on ASR, by at most 5\% WER. Overall, for supervised tasks (ASR and MT), the plain zero-shot models have reasonable performance.
	
	However, the zero-shot models are unable to perform zero-shot ST. We observe that all zero-shot predictions are in the wrong language. The models always output \textit{EN text} for \textit{EN audio} input, even when we set the target-language token to be \textit{$<$DE$>$}. A possible reason is as follows. In the training data, \textit{EN audio} input always has \textit{EN text} output and \textit{EN text} input always has \textit{DE text} output. Since the representation of audio and text are different, the model ignores the target-language tokens and decides on the output language based on the modality of the input (audio or text), hence unable to perform zero-shot. One of our experiment results supports this hypothesis, which will be discussed in detail in Section \ref{sec:ad4_exp}.
	
	\begin{table}[htbp]
		\caption{Plain zero-shot models versus single-task models.}
		\begin{center}
			\begin{tabular}{|r|l|r|r|r|}
				\hline
				\multicolumn{1}{|c|}{\begin{tabular}[c]{@{}c@{}}Data portion\end{tabular}} &
				\multicolumn{1}{c|}{Model type} &
				\multicolumn{1}{c|}{ASR} &
				\multicolumn{1}{c|}{MT} &
				\multicolumn{1}{c|}{Zero-shot ST} \\ \hline
				25\%  & Single-task                                                           & 43.6 & 23.1 & --                                                              \\ \hline
				25\%  & \begin{tabular}[c]{@{}l@{}}Plain zero-shot (ZS)\end{tabular} & 48.1 & 23.5 & \begin{tabular}[c]{@{}r@{}}0.3\end{tabular} \\ \hline \hline
				100\% & Single-task                                                           & 25.8 & 33.0 & --                                                              \\ \hline
				100\% & \begin{tabular}[c]{@{}l@{}}Plain zero-shot (ZS) \end{tabular} & 28.4 & 32.8 & \begin{tabular}[c]{@{}r@{}}0.6\end{tabular} \\ \hline
			\end{tabular}
			\label{tab:baseZS}
		\end{center}
	\end{table}
	
	We build few-shot ST models by fine-tuning the plain zero-shot model with 10\% of ST data. We compare the few-shot models to some baselines: direct end-to-end models trained on 10\% of ST data from scratch and ST models fine-tuned from ASR with 10\% of ST data. We observe that direct end-to-end ST models have poor performance, at 0.5 BLEU points. On the other hand, the fine-tuned ASR models have better performance, hence we compare them to our few-shot models as shown in Table \ref{tab:plain_ft}. Our models fine-tuned from plain zero-shot outperform the models fine-tuned from ASR. When the amount of pre-training data (i.e., data that the models were trained on before fine-tuning) increases, the performance gap becomes more significant: from +0.3 to +1.4 BLEU points.
	
	\begin{table}[htbp]
		\caption{Few-shot models fine-tuned from plain zero-shot and ASR.}
		\begin{center}
			\begin{tabular}{lllll}
				\hline
				\multicolumn{1}{|c|}{\begin{tabular}[c]{@{}c@{}}Pre-training \\ data portion\end{tabular}} &
				\multicolumn{1}{|c|}{\begin{tabular}[c]{@{}c@{}}Pre-trained model type\end{tabular}} &
				\multicolumn{1}{c|}{\begin{tabular}[c]{@{}c@{}}Fine-tuning data\end{tabular}} &
				\multicolumn{2}{c|}{ST score} \\ \hline
				\multicolumn{1}{|l|}{25\%} &
				\multicolumn{1}{l|}{ASR} &
				\multicolumn{1}{l|}{10\% ST} &
				\multicolumn{2}{l|}{3.7} \\ \hline
				\multicolumn{1}{|l|}{25\%} &
				\multicolumn{1}{l|}{Plain zero-shot (ZS)} &
				\multicolumn{1}{l|}{10\% ST} &
				\multicolumn{2}{l|}{4.0 \textbf{(+0.3)}} \\ \hline \hline
				\multicolumn{1}{|l|}{100\%} &
				\multicolumn{1}{l|}{ASR} &
				\multicolumn{1}{l|}{10\% ST} &
				\multicolumn{2}{l|}{8.4} \\ \hline
				\multicolumn{1}{|l|}{100\%} &
				\multicolumn{1}{l|}{Plain zero-shot (ZS)} &
				\multicolumn{1}{l|}{10\% ST} &
				\multicolumn{2}{l|}{9.8 \textbf{(+1.4)}} \\ \hline
				\multicolumn{5}{l}{\begin{tabular}[c]{@{}l@{}}Numbers in the brackets are comparison to the fine-tuned ASR models.\end{tabular}}
			\end{tabular}
			\label{tab:plain_ft}
		\end{center}
	\end{table}
	
	In the next experiments, we compare the performance of few-shot models using additional approaches to the few-shot models using the plain setting (instead of comparing them to the direct end-to-end ST and fine-tuned ASR models) to see the effect of the additional approaches.
	
	\subsection{Disentangling Positional Information}
	For the zero-shot model with residual connections in the middle layer of the shared encoder part removed, we observe that the model failed to learn the ASR task. The validation loss for the ASR task could not converge, and the WER score on the test set is 102.2\% (i.e., the model output transcriptions are mostly wrong). However, for the MT task, the validation loss can converge, and the BLEU score on the test set is 13.7. This is worse compared to the plain zero-shot model (with 23.1 BLEU points on MT task), but it still proves that MT task can be learned with the middle residual connections removed. When performing zero-shot ST, the model still outputs the wrong language (\textit{EN} instead of \textit{DE}).
	
\begin{figure}[htbp]
		\centerline{\includegraphics[width=0.73\linewidth]{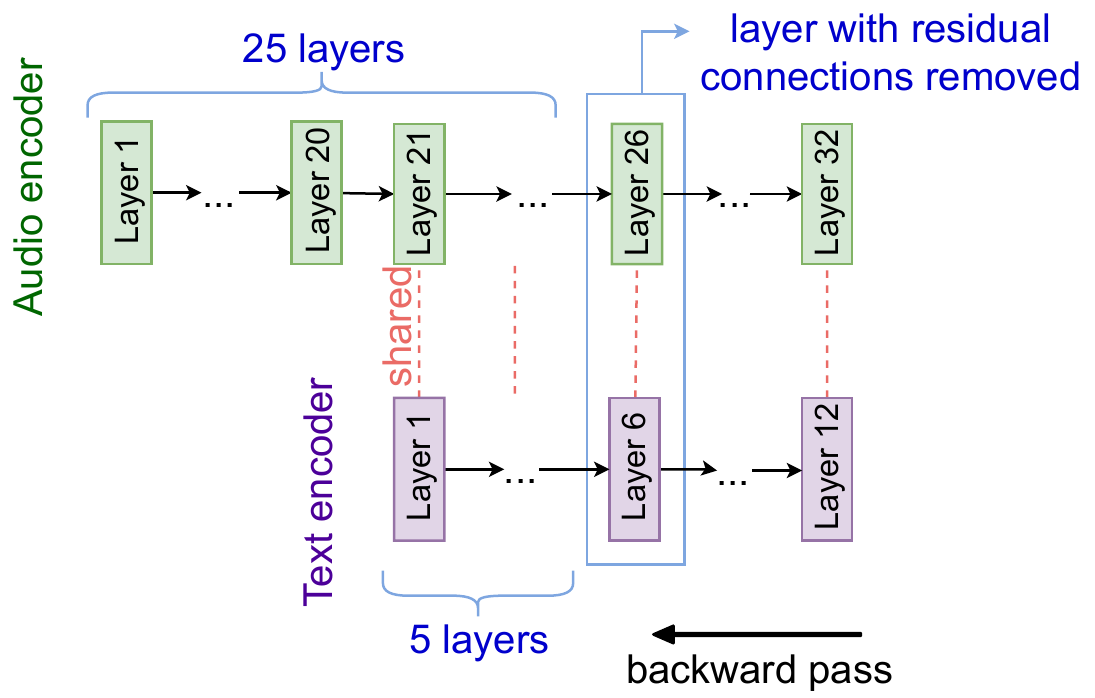}}
		\caption{Illustration of the Disentangling Positional Information approach.}
		\label{fig:DEPI}
	\end{figure}

	
	The depth of the audio encoder is believed to be the reason why the model cannot learn the ASR task when we apply this approach. As described in Section \ref{sec:DEPI}, we remove the residual connections of the middle text encoder layer, which is shared with the audio encoder. In Fig. \ref{fig:DEPI}, for the audio encoder, in the backward pass, there are 25 layers below the residual-connection-removed layer. Therefore, these 25 layers are learnt less, which makes it difficult to learn the ASR task. For the text encoder, there are only 5 layers below the residual-connection-removed layer. Thus, the model can still learn the MT task, although the performance is worse than the usual setting. The difference between text and audio input is also a part of the problem. The idea of this approach was to relax the strong positional correspondence of the output to input tokens. Audio input consists of feature frames, while text input consists of subwords. Unlike the subwords, the feature frames are highly correlated to each other; e.g., some neighboring frames might belong to the same spoken phoneme. Therefore, the position of the frames is important, and it is difficult for the upper 6 layers of the audio encoder to capture the necessary positional information. In the original paper \cite{DEPI}, this approach was applied on zero-shot multilingual MT (involving only text) with a smaller number of encoder layers, hence it worked on their setting but did not work in our case.
	
	\subsection{Auxiliary loss function} \label{sec:auxloss_exp}
	Table \ref{tab:auxloss} shows the results of using an auxiliary loss to minimize the difference between text and audio encoder output in comparison with no auxiliary loss. As can be seen, with different weights of the auxiliary loss, the BLEU scores for zero-shot ST remain very low (all under 0.4 when training on 25\% of the data and 0.65 when training on full data), meaning that the models are still unable to perform zero-shot ST. We observe that all models in this experiment output the wrong language (\textit{EN} instead of \textit{DE}) when performing zero-shot ST.
	
	\begin{table}[htbp]
		\caption{Zero-shot models with auxiliary loss.}
		\begin{center}
			\begin{tabular}{|r|r|r|r|r|}
				\hline
				\multicolumn{1}{|c|}{\begin{tabular}[c]{@{}c@{}}Data portion\end{tabular}} &
				\multicolumn{1}{c|}{\begin{tabular}[c]{@{}c@{}}Aux. loss weight\end{tabular}} &
				\multicolumn{1}{c|}{ASR} &
				\multicolumn{1}{c|}{MT} &
				\multicolumn{1}{c|}{Zero-shot ST} \\ \hline
				25\%  & 0.0 & 48.1 & 23.5 & 0.32 \\ \hline
				25\%  & 0.1 & 45.7 & 22.7 & 0.38 \\ \hline
				25\%  & 1.0 & 45.7 & 22.5 & 0.38 \\ \hline
				25\%  & 5.0 & 43.9 & 21.9 & 0.39 \\ \hline \hline
				100\% & 0.0 & 28.4 & 32.8 & 0.63 \\ \hline
				100\% & 5.0 & 26.9 & 32.5 & 0.65 \\ \hline
			\end{tabular}
			\label{tab:auxloss}
		\end{center}
	\end{table}
	
	Recall that our metric for encoder output difference is the squared error of mean-pool over time. We observe that the number of time steps of an \textit{EN audio} sample is much higher than that of the aligned \textit{EN text} sample, since audio input contains many audio frames while text input consists of subwords. When we calculate the squared error of mean-pool over time, i.e., averaging over the time steps, the difference in the number of time steps is not considered. This possibly explains why the representation of audio and text remains different, thus the model cannot perform zero-shot ST.
	
	When training on 25\% of the data, we observe that the models with higher auxiliary loss weight have better ASR performance. One possible reason is that, the auxiliary loss function helps the models to represent text and audio more similarly (although still not enough to perform zero-shot ST), hence \textit{EN audio} and \textit{EN text} become more similar, making it easier to transcribe \textit{EN audio} $\rightarrow$ \textit{EN text}. The model using auxiliary loss with weight 5 outperforms the plain model on the ASR task by $-4.2$\% WER. When training on full data, the performance gap is no longer as large, at only $-1.5$\% WER. The reason is that it is easier for the model to learn the ASR task with a large amount of data, hence the help of the auxiliary loss is no longer as significant.
	
	Since we want the audio and text representation to be as similar as possible to encourage zero-shot, we use the weight of 5, which is the highest weight we have experimented with, whenever the auxiliary loss is used in the next experiments.
	
	We build few-shot ST models by fine-tuning the zero-shot model using auxiliary loss with 10\% of ST data. We compare this to the plain setting, as shown in Table \ref{tab:auxloss_ft}. Observe that the auxiliary loss helps improving the ST performance of the few-shot models: the BLEU score increased by +0.3 with 25\% pre-training data and +0.8 with full pre-training data.
	\subsection{Data augmentation} \label{sec:ad4_exp}
	\begin{table}[htbp]
		\caption{Few-shot models from setting with auxiliary loss.}
		\begin{center}
			\begin{tabular}{rlll}
				\hline
				\multicolumn{1}{|c|}{\begin{tabular}[c]{@{}c@{}}Pre-training \\ data portion\end{tabular}} &
				\multicolumn{1}{|c|}{\begin{tabular}[c]{@{}c@{}}Pre-trained \\ model type\end{tabular}} &
				\multicolumn{1}{c|}{\begin{tabular}[c]{@{}c@{}}Fine-tuning data\end{tabular}} &
				\multicolumn{1}{c|}{ST score} \\ \hline
				\multicolumn{1}{|r|}{25\%} &
				\multicolumn{1}{l|}{\begin{tabular}[c]{@{}l@{}}ZS + auxiliary loss\end{tabular}} &
				\multicolumn{1}{l|}{10\% ST} &
				\multicolumn{1}{l|}{$\>$ 4.3 \textbf{(+0.3)}} \\ \hline 
				\multicolumn{1}{|r|}{100\%} &
				\multicolumn{1}{l|}{\begin{tabular}[c]{@{}l@{}}ZS + auxiliary loss\end{tabular}} &
				\multicolumn{1}{l|}{10\% ST} &
				\multicolumn{1}{l|}{10.6 \textbf{(+0.8)}} \\ \hline
				\multicolumn{4}{l}{\begin{tabular}[c]{@{}l@{}}Numbers in the brackets are comparison to the corresponding plain setting.\end{tabular}}
			\end{tabular}
			\label{tab:auxloss_ft}
		\end{center}
	\end{table}
	
\begin{figure}[htbp]
		\centerline{\includegraphics[width=0.52\textwidth]{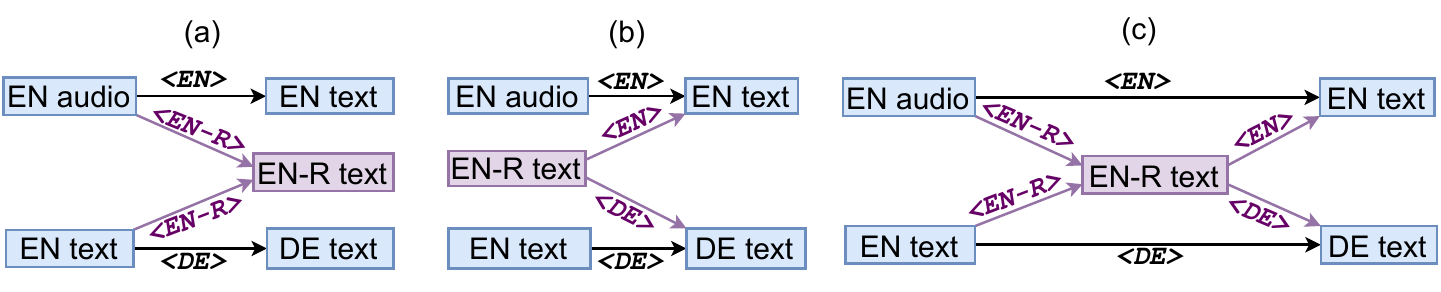}}
		\caption{Different settings for data augmentation. The black arrows are the main training directions. The purple arrows are the artificial directions.}
		\label{fig:EnDeAd}
	\end{figure}

	We experiment with different sets of augmented training data as described in Fig. \ref{fig:EnDeAd}. The amount of artificial data is half of the amount of the main data. In setting (a), both \textit{EN audio} and \textit{EN text} input has two target languages. In setting (b), we have the same \textit{EN-R text} input with \textit{EN} and \textit{DE} output so that the model can learn to switch between \textit{EN} and \textit{DE} targets. In setting (c), we include the artificial data of both (a) and (b). 
	
	The results of models trained on 25\% of the data are shown in Table \ref{tab:ad4_25} and Table \ref{tab:ad4-st-25}. From Table \ref{tab:ad4_25}, it can be seen that models with augmented training data are unable to perform zero-shot ST task. The BLEU scores are very low, where the highest is only at 0.45. However, we observe that the models output some tokens in the correct language (\textit{DE}) when performing zero-shot ST, which was not the case in the plain setting. Therefore, we report the percentage of tokens in each language, shown in Table \ref{tab:ad4-st-25}, to further examine the zero-shot ST output. Note that we do not take the \textit{EN-R} language into account, since all models rarely output \textit{EN-R} tokens while doing zero-shot ST. We report the percentages of tokens belonging to both \textit{EN} and \textit{DE}, tokens belonging to only \textit{EN} and tokens belonging to only \textit{DE}. Since the target language is \textit{DE}, the more tokens output in \textit{DE} the better. Looking at Table \ref{tab:ad4-st-25}, the models with augmented training data output more \textit{DE}-only tokens than the plain model. The model with setting (c) has significantly more \textit{DE}-only tokens than the other, at 3.209\%. Although this is far from the ground truth (53.485\%), it still indicates that the models are starting to learn the target-language tokens using augmented training data. We use the most promising setting, i.e., setting (c), whenever data augmentation is used in the subsequent experiments.
	
	\begin{table}[htbp]
		\caption{Zero-shot models trained on 25\% data with augmentation.}
		\begin{center}
			\begin{tabular}{|r|l|r|r|r|}
				\hline
				\multicolumn{1}{|c|}{\begin{tabular}[c]{@{}c@{}}Data portion\end{tabular}} &
				Model type &
				\multicolumn{1}{l|}{ASR} &
				\multicolumn{1}{l|}{MT} &
				\multicolumn{1}{l|}{Zero-shot ST} \\ \hline
				25\% & Plain zero-shot (ZS)        & 48.1 & 23.5 & 0.32 \\ \hline
				25\% & ZS + augmented data (a) & 47.9 & 24.1 & 0.37 \\ \hline
				25\% & ZS + augmented data (b) & 51.6 & 22.7 & 0.34 \\ \hline
				25\% & ZS + augmented data (c) & 47.3 & 23.2 & 0.45 \\ \hline
			\end{tabular}
			\label{tab:ad4_25}
		\end{center}
	\end{table}
	
	\begin{table}[htbp]
		\caption{Percentage of output tokens in each language of models trained on 25\% of the data.}
		\begin{center}
			\begin{tabular}{rlrrr}
				\hline
				\multicolumn{1}{|c|}{\begin{tabular}[c]{@{}c@{}}Data portion\end{tabular}} &
				\multicolumn{1}{l|}{Model type} &
				\multicolumn{1}{c|}{EN $\cap$ DE} &
				\multicolumn{1}{c|}{EN} &
				\multicolumn{1}{c|}{DE} \\ \hline
				\multicolumn{1}{|r|}{25\%} & \multicolumn{1}{l|}{Plain zero-shot (ZS)}         & \multicolumn{1}{r|}{55.380} & \multicolumn{1}{r|}{44.618} & \multicolumn{1}{r|}{0.002} \\ \hline
				\multicolumn{1}{|r|}{25\%} & \multicolumn{1}{l|}{ZS + augmented data (a)} & \multicolumn{1}{r|}{55.095} & \multicolumn{1}{r|}{44.899} & \multicolumn{1}{r|}{0.004} \\ \hline
				\multicolumn{1}{|r|}{25\%} & \multicolumn{1}{l|}{ZS + augmented data (b)} & \multicolumn{1}{r|}{60.381} & \multicolumn{1}{r|}{39.345} & \multicolumn{1}{r|}{0.274} \\ \hline
				\multicolumn{1}{|r|}{25\%} &
				\multicolumn{1}{l|}{ZS + augmented data (c)} &
				\multicolumn{1}{r|}{\textbf{63.276}} &
				\multicolumn{1}{r|}{\textbf{33.491}} &
				\multicolumn{1}{r|}{\textbf{3.209}} \\ \hline \hline
				\multicolumn{1}{|r|}{25\%} &
				\multicolumn{1}{l|}{\begin{tabular}[c]{@{}l@{}}Human-labeled\end{tabular}} &
				\multicolumn{1}{r|}{46.512} &
				\multicolumn{1}{r|}{0.000} &
				\multicolumn{1}{r|}{53.485} \\ \hline
				\multicolumn{5}{l}{ENR's statistics are excluded.}                                                                                                                
			\end{tabular}
			\label{tab:ad4-st-25}
		\end{center}
	\end{table}
	
	Another interesting observation from Table \ref{tab:ad4-st-25} is that the percentages of tokens belonging to both \textit{EN} and \textit{DE} of the output zero-shot ST (all above 55\%) are higher than that of the human-labeled translation (at 46.512\%). The zero-shot models seem to output more of these tokens to commit to what they have observed a lot during training (\textit{EN audio} $\rightarrow$ \textit{EN text}) and satisfy the \textit{$<$DE$>$} target-language token at the same time.
	
	The result above (model being able to output some tokens in correct language) supports our former hypothesis in Section \ref{sec:baseZS}: plain zero-shot models output the wrong language due to the difference in text-audio representation. In the plain setting, the training samples are \textit{EN audio} $\rightarrow$ \textit{EN text} and \textit{EN text} $\rightarrow$ \textit{DE text}. We stated that, due to the difference between text and audio, the models decide on the output language based on the modality, i.e., output \textit{EN} for all audio input and output \textit{DE} for all text input. With data augmentation, both \textit{EN text} and \textit{EN audio} input have more than one target output language during training, hence the model is forced to learn the target-language token in order to decide which language to output.
	
	We train the most promising setting, i.e., setting (c), on the full data. The results are shown in Table \ref{tab:ad4_full} and Table \ref{tab:ad4-st-full}. Table \ref{tab:ad4_full} shows that this method does not scale well. When being trained on full data, the model with augmented data (c) has a significantly worse performance on ASR (with 47.6\% WER) compared to the plain model (with 28.4\% WER), while the performance on MT task stays approximately the same. This was not the case when the model was trained on 25\% of the data. We suspect that, since there are 4 text-related tasks and only 2 audio-related tasks during training, the model focus more on the text-related tasks (MT tasks) than audio-related task (including ASR task). This did not happen when model was trained on 25\% of the data since with 25\% of the data, the ASR task cannot be learned any better, therefore the performance gap was not significant. 
	
	To overcome the shortcoming of the data-augmentation approach when training on full data, we fine-tune the plain-zero shot model with the augmented data instead of training on it from scratch. Since the plain model has good performance on ASR, we expect it to be a good starting point to achieve decent performance using augmented data in the end. This is indeed the case, as shown in Table \ref{tab:ad4_full}. The fine-tuned model has similar performance on supervised tasks compared to the plain zero-shot model. However, the fine-tuned model cannot output as many \textit{DE} tokens as the model trained with augmented data from scratch, as shown in Table \ref{tab:ad4-st-full}. 
	
	\begin{table}[htbp]
		\caption{Zero-shot models trained on full data with augmentation.}
		\begin{center}
			\begin{tabular}{|r|l|r|r|r|}
				\hline
				\multicolumn{1}{|c|}{\begin{tabular}[c]{@{}c@{}}Data portion\end{tabular}} &
				Model type &
				\multicolumn{1}{l|}{ASR} &
				\multicolumn{1}{l|}{MT} &
				\multicolumn{1}{l|}{Zero-shot ST} \\ \hline
				100\% & Plain zero-shot (ZS)        & 28.4 & 32.8 & 0.63 \\ \hline
				100\% & ZS + augmented data (c) & 47.6 & 31.6 & 0.53 \\ \hline
				100\% &
				\begin{tabular}[c]{@{}l@{}}ZS + augmented data (c)\\ (fine-tuned from plain)\end{tabular} &
				27.6 &
				32.2 &
				0.68 \\ \hline
			\end{tabular}
			\label{tab:ad4_full}
		\end{center}
	\end{table}
	
	\begin{table}[htbp]
		\caption{Percentage of output tokens in each language of models trained on full data.}
		\begin{center}
			\begin{tabular}{rlrrr}
				\hline
				\multicolumn{1}{|c|}{\begin{tabular}[c]{@{}c@{}}Data portion\end{tabular}} &
				\multicolumn{1}{l|}{Model type} &
				\multicolumn{1}{c|}{EN $\cap$ DE} &
				\multicolumn{1}{c|}{EN} &
				\multicolumn{1}{c|}{DE} \\ \hline
				\multicolumn{1}{|r|}{100\%} & \multicolumn{1}{l|}{Plain zero-shot (ZS)}         & \multicolumn{1}{r|}{54.203} & \multicolumn{1}{r|}{45.797} & \multicolumn{1}{r|}{0.000} \\ \hline
				\multicolumn{1}{|r|}{100\%} & \multicolumn{1}{l|}{ZS + augmented data (c)} & \multicolumn{1}{r|}{63.432} & \multicolumn{1}{r|}{35.024} & \multicolumn{1}{r|}{1.538} \\ \hline
				\multicolumn{1}{|r|}{100\%} &
				\multicolumn{1}{l|}{\begin{tabular}[c]{@{}l@{}}ZS + augmented data (c)\\ (fine-tuned from plain)\end{tabular}} &
				\multicolumn{1}{r|}{55.700} &
				\multicolumn{1}{r|}{44.251} &
				\multicolumn{1}{r|}{0.038} \\ \hline \hline
				\multicolumn{1}{|r|}{100\%} &
				\multicolumn{1}{l|}{\begin{tabular}[c]{@{}l@{}}Human-labeled\end{tabular}} &
				\multicolumn{1}{r|}{45.558} &
				\multicolumn{1}{r|}{0.000} &
				\multicolumn{1}{r|}{54.441} \\ \hline
				\multicolumn{5}{l}{ENR's statistics are excluded.}           
			\end{tabular}
			\label{tab:ad4-st-full}
		\end{center}
	\end{table}
	
	We then build few-shot ST models by fine-tuning the zero-shot model trained on augmented data with 10\% of ST data. We compare this to the plain setting, as shown in Table \ref{tab:ad4_ft}. The few-shot model using the data augmentation approach also does not scale well: the BLEU score decreases by 4.4 when the pre-training data are large, which is not the case when the pre-training data are small. This is expected, since the model trained on full augmented data from scratch has poor ASR performance, hence it is a worse baseline for fine-tuning. In contrast, the few-shot model using the augmented-data approach fine-tuned from the plain model has a significant improvement. The BLEU score for ST task increases by +1.7, which is more than twice larger than the gap of models pre-trained of 25\% of the data, at +0.8 BLEU points.
	
	\begin{table}[htbp]
		\caption{Few-shot models from the setting with augmented data (c).}
		\begin{center}
			\begin{tabular}{rllll}
				\hline
				\multicolumn{1}{|c|}{\begin{tabular}[c]{@{}c@{}}Pre-training \\ data portion\end{tabular}} &
				\multicolumn{1}{|c|}{\begin{tabular}[c]{@{}c@{}}Pre-trained \\ model type\end{tabular}} &
				\multicolumn{1}{c|}{\begin{tabular}[c]{@{}c@{}}Fine-tuning data\end{tabular}} &
				\multicolumn{2}{c|}{ST score} \\ \hline
				\multicolumn{1}{|r|}{25\%} &
				\multicolumn{1}{l|}{\begin{tabular}[c]{@{}l@{}}ZS + augmented data\end{tabular}} &
				\multicolumn{1}{l|}{10\% ST} &
				\multicolumn{2}{l|}{\text{ }4.8 $\-$ \textbf{(+0.8)}} \\ \hline 
				\multicolumn{1}{|r|}{100\%} &
				\multicolumn{1}{l|}{\begin{tabular}[c]{@{}l@{}}ZS + augmented data\end{tabular}} &
				\multicolumn{1}{l|}{10\% ST} &
				\multicolumn{2}{l|}{\text{ }5.4 ($-4.4$)} \\ \hline
				\multicolumn{1}{|r|}{100\%} &
				\multicolumn{1}{l|}{\begin{tabular}[c]{@{}l@{}}ZS + augmented data \\ (fine-tuned from plain)\end{tabular}} &
				\multicolumn{1}{l|}{10\% ST} &
				\multicolumn{2}{l|}{11.5 $\-$ \textbf{(+1.7)}} \\ \hline
				\multicolumn{5}{l}{\begin{tabular}[c]{@{}l@{}}Numbers in the brackets are comparison to the corresponding plain setting.\end{tabular}}
			\end{tabular}
			\label{tab:ad4_ft}
		\end{center}
	\end{table}
	
	\subsection{Additional opposite training data} 
	In this experiment, we train the model with the main ASR and MT data of \textit{EN} $\rightarrow$ \textit{DE} direction and the additional ASR and MT data of the \textit{DE} $\rightarrow$ \textit{EN} direction. 
	The amount of additional data is half of the amount of the main training data.
	
	The results are shown in Table \ref{tab:bi}. The BLEU scores of zero-shot ST by the models trained on additional opposite data are still low (0.47 when trained on 25\% of the data and 1.36 when trained on full data). We observe that the zero-shot ST output by the models is now mostly in the correct language - \textit{DE}. However, in many sentences, it looks like the models are trying to recognize \textit{DE text} from \textit{EN audio}, instead of outputting the \textit{DE} translation from \textit{EN audio} as we hope.
	
	We believe the text-audio difference is the reason why our models with additional opposite data have poor performance on zero-shot ST, although they are able to output the correct language. Our models try to recognize \textit{DE text} from \textit{EN audio}, meaning that they are using the \textit{DE} ASR task that was presented during training, instead of using what they have learnt from the MT tasks as well. Because of the text-audio difference, the models learn the ASR and MT tasks in very different ways, hence it is difficult to do zero-shot prediction using the knowledge from both the ASR and MT tasks.
	
	\begin{table}[htbp]
		\caption{Zero-shot models with opposite data.}
		\begin{center}
			\begin{tabular}{|r|l|r|r|r|}
				\hline
				\multicolumn{1}{|c|}{\begin{tabular}[c]{@{}c@{}}Data portion\end{tabular}} &
				Model type &
				\multicolumn{1}{l|}{ASR} &
				\multicolumn{1}{l|}{MT} &
				\multicolumn{1}{l|}{Zero-shot ST} \\ \hline
				25\%  & Plain zero-shot (ZS)   & 48.1 & 23.5 & 0.32 \\ \hline
				25\%  & ZS + opposite data & 48.8 & 24.5 & 0.47 \\ \hline \hline
				100\% & Plain zero-shot (ZS)   & 28.4 & 32.8 & 0.63 \\ \hline
				100\% & ZS + opposite data & 26.8 & 32.6 & 1.36 \\ \hline
			\end{tabular}
			\label{tab:bi}
		\end{center}
	\end{table}
	
	We then build few-shot ST models by fine-tuning the zero-shot model using the additional opposite data approach with 10\% of ST data. We compare this to the plain setting, as shown in Table \ref{tab:bi_ft}. Observe that this approach helps improving the ST performance of the few-shot models: the BLEU score increased by +0.8 when the pre-training data portion is 25\% and +0.5 when the pre-training data portion is full.
	
	\begin{table}[htbp]
		\caption{Few-shot models from setting with opposite data.}
		\begin{center}
			\begin{tabular}{rlll}
				\hline
				\multicolumn{1}{|c|}{\begin{tabular}[c]{@{}c@{}}Pre-training \\ data portion\end{tabular}} &
				\multicolumn{1}{|c|}{\begin{tabular}[c]{@{}c@{}}Pre-trained \\ model type\end{tabular}} &
				\multicolumn{1}{c|}{\begin{tabular}[c]{@{}c@{}}Fine-tuning data\end{tabular}} &
				\multicolumn{1}{c|}{ST score} \\ \hline
				\multicolumn{1}{|r|}{25\%} &
				\multicolumn{1}{l|}{\begin{tabular}[c]{@{}l@{}}ZS + opposite data\end{tabular}} &
				\multicolumn{1}{l|}{10\% ST} &
				\multicolumn{1}{l|}{\text{ }4.8 \textbf{(+0.8)}} \\ \hline 
				\multicolumn{1}{|r|}{100\%} &
				\multicolumn{1}{l|}{\begin{tabular}[c]{@{}l@{}}ZS + opposite data\end{tabular}} &
				\multicolumn{1}{l|}{10\% ST} &
				\multicolumn{1}{l|}{10.3 \textbf{(+0.5)}} \\ \hline
				\multicolumn{4}{l}{\begin{tabular}[c]{@{}l@{}}Numbers in the brackets are comparison to the corresponding plain setting.\end{tabular}}
			\end{tabular}
			\label{tab:bi_ft}
		\end{center}
	\end{table}
	
	\subsection{Combination of approaches} \label{sec:combine_exp}
	In the previous experiments, we observe that using auxiliary loss, augmented data and additional opposite data are the most promising approaches. Hence, we experiment with the combinations of these approaches.
	
	We first combine data augmentation with auxiliary loss. Section \ref{sec:ad4_exp} shows that the data augmentation approach does not scale well and only gives decent performance when being fine-tuned from the plain model on full data. Hence, we combine this with the auxiliary loss approach by including the auxiliary loss when fine-tuning. We observe that the performance of the zero-shot model with the combined auxiliary loss is almost the same as without auxiliary loss. The score differences are at most 0.1 for all the ASR, MT and ST tasks. When building the corresponding few-shot model using 10\% of ST data, we also observe no difference in the ST performance with and without auxiliary loss: the scores are both 11.5 BLEU points. For the detailed experiment results, see Appendix A.

	%

	Next, we combine the additional opposite training data and auxiliary loss approaches. Table \ref{tab:bi_aux} shows the experiment results of zero-shot models with this combination compared to the plain models. Observe that the zero-shot models with this combined approach have the same or better performance on the supervised tasks (ASR and MT). 
	For zero-shot ST, this is the best result so far, at 0.72 BLEU points with 25\% training data and 1.51 BLEU points with full training data, although more work still needs to be done until zero-shot ST can be put to practical use. This combined approach also improves the performance of few-shot models, as can be seen in Table \ref{tab:bi_aux_ft}. The BLEU score increased by +1.7 with 25\% pre-training data and +2.5 with full pre-training data.
	
	\begin{table}[htbp]
		\caption{Zero-shot models with opposite data and auxiliary loss.}
		\begin{center}
			\begin{tabular}{|r|l|r|r|r|}
				\hline
				\multicolumn{1}{|c|}{\begin{tabular}[c]{@{}c@{}}Data portion\end{tabular}} &
				Model type &
				\multicolumn{1}{l|}{ASR} &
				\multicolumn{1}{l|}{MT} &
				\multicolumn{1}{l|}{\begin{tabular}[c]{@{}c@{}}Zero-shot \\ST\end{tabular}} \\ \hline
				25\%  & Plain zero-shot (ZS)                                                              & 48.1 & 23.5 & 0.32 \\ \hline
				25\%  & \begin{tabular}[c]{@{}l@{}}ZS + opposite data + aux. loss\end{tabular} & 39.4 & 24.1 & 0.72 \\ \hline \hline
				100\% & Plain zero-shot (ZS)                                                               & 28.4 & 32.8 & 0.63 \\ \hline
				100\% & \begin{tabular}[c]{@{}l@{}}ZS + opposite data + aux. loss\end{tabular} &  25.4    &   32.8   &   1.51   \\ \hline
			\end{tabular}
			\label{tab:bi_aux}
		\end{center}
	\end{table}
	
	\begin{table}[htbp]
		\caption{Few-shot models using opposite data and auxiliary loss.}
		\begin{center}
			\begin{tabular}{rllll}
				\hline
				\multicolumn{1}{|c|}{\begin{tabular}[c]{@{}c@{}}Pre-training \\ data portion\end{tabular}} &
				\multicolumn{1}{|c|}{\begin{tabular}[c]{@{}c@{}}Pre-trained \\ model type\end{tabular}} &
				\multicolumn{1}{c|}{\begin{tabular}[c]{@{}c@{}}Fine-tuning \\data\end{tabular}} &
				\multicolumn{2}{c|}{ST score} \\ \hline
				\multicolumn{1}{|r|}{25\%} &
				\multicolumn{1}{l|}{\begin{tabular}[c]{@{}l@{}}ZS + opposite data + aux. loss\end{tabular}} &
				\multicolumn{1}{l|}{10\% ST} &
				\multicolumn{2}{l|}{\text{ }5.7 \textbf{(+1.7)}} \\ \hline
				\multicolumn{1}{|r|}{100\%} &
				\multicolumn{1}{l|}{\begin{tabular}[c]{@{}l@{}}ZS + opposite data + aux. loss\end{tabular}} &
				\multicolumn{1}{l|}{10\% ST} &
				\multicolumn{2}{l|}{12.3 \textbf{(+2.5)}} \\ \hline 
				\multicolumn{5}{l}{\begin{tabular}[c]{@{}l@{}}Numbers in the brackets are comparison to the corresponding plain setting.\end{tabular}}
			\end{tabular}
			\label{tab:bi_aux_ft}
		\end{center}
	\end{table}
	
	From this experiment, we see that auxiliary loss helps improving the ST performance when combined with the opposite data setting, but not with the data augmentation setting. We will further investigate this phenomenon in Section \ref{sec:analysis}.
	
	\subsection{More on few-shot Speech Translation}
	In the previous experiments, we have seen that although our approaches do not enable zero-shot ST to the point where it can be used practically, they are useful to improve the ST performance when a limited amount of ST data is available. Considering each approach individually, data augmentation is the best approach, where the few-shot model's ST performance is 4.8 BLEU points with 25\% pre-training data, and 11.5 BLEU points with full pre-training data (with full training data, we need to fine-tune from the plain model). The approach using opposite data has the same ST performance as using augmented data for few-shot models pre-trained on 25\% of the data. However, unlike the opposite data approach, the data augmentation approach does not require any real data corpus in addition to the main ASR and MT data, hence data augmentation would be a better choice. Considering the approaches in combination, we see that adding opposite training data combined with auxiliary loss is the best, where the few-shot model's ST performance is 5.7 BLEU points with 25\% pre-training data, and 12.3 BLEU points with full pre-training data. Comparing this best few-shot model to the direct end-to-end model trained on the same limited amount of ST data (with ST performance of 0.5 BLEU points), we see an improvement of +5.2 BLEU points when the few-shot model is pre-trained on 25\% of the data and +11.8 BLEU points when the few-shot model is pre-trained on full data. We also compare our best few-shot model to the more competitive baseline (i.e., ASR model fine-tuned with the same amount of ST data) and observe an improvement of +2.0 BLEU points when the models are pre-trained on 25\% data and +3.9 BLEU points when the models are pre-trained on full data.

	We also experiment with a higher amount of ST data for fine-tuning (25\% instead of 10\%). We observe that, with more ST data, the overall ST performance improves, but the gain from additional approaches compared to the plain setting is less significant. We also see that our best few-shot model using opposite data and auxiliary loss has 14.2 BLEU points using 25\% of the ST data, which is only 0.7 points lower than that of the direct end-to-end ST model using full ST data (at 14.9). For the detailed experiment result, see Appendix B.

	We observe that, when fine-tuning a zero-shot model with only ST data, the model forgets the tasks that it was previously trained on. The model can no longer perform ASR (it output \textit{DE text}, while ASR requires \textit{EN text} output), and the MT performance is significantly worse. This problem can be solved by fine-tuning the zero-shot model with small amounts of all ASR, MT and ST data. We observe this new few-shot model is a middle ground: it has reasonable performance for all three tasks, but the ASR and MT performance is slightly worse than the model before fine-tuning and the ST performance is slightly worse than the model fine-tuned with only ST data. See Appendix C for the detailed experiment result.

	\section{Analysis} \label{sec:analysis}
	In this section, we further analyze our models from Section \ref{sec:results}. We showed that text-audio difference is the main challenge for zero-shot ST. Thus, we investigate the similarity between aligned text and audio mean-pooled encoder output using Singular Vector Canonical Correlation Analysis (SVCCA) \cite{NIPS2017_7188}. Higher SVCCA scores mean more text-audio similarity, i.e., more modality-independent representation. We use the score of the plain setting as the baseline for comparison. 
	
\begin{figure}[htbp]
		\centerline{\includegraphics[width=0.46\textwidth]{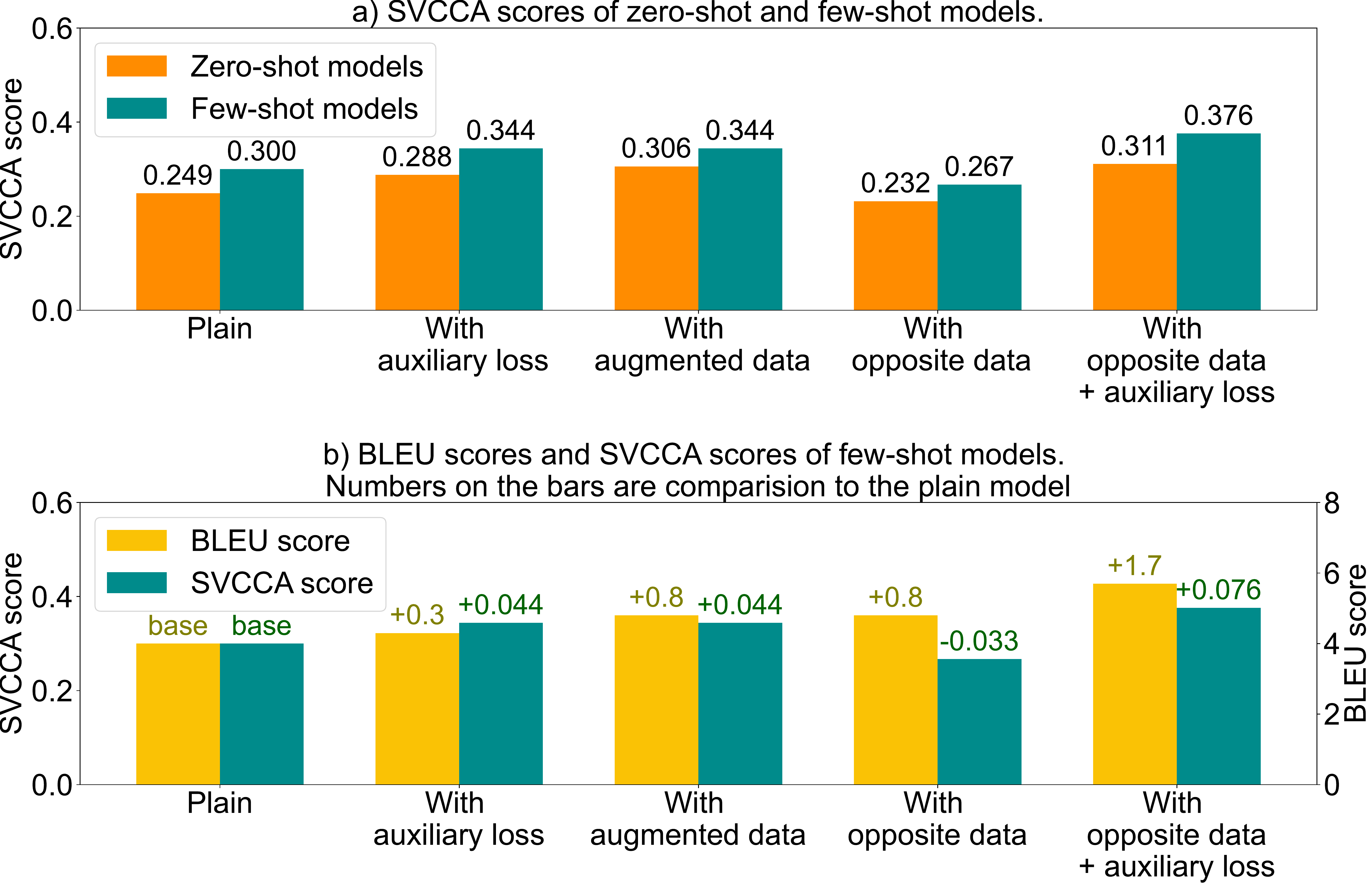}}
		\caption{Analysis of models pre-trained on 25\% of the data.}
		\label{fig:models25}
	\end{figure}

	The analysis of models trained on 25\% of the data is shown in Fig. \ref{fig:models25}. Looking at Fig. \ref{fig:models25}a, all of our approaches help increasing text-audio similarity, except adding opposite training data. Trivially, the models with auxiliary loss have higher SVCCA scores, since we designed our auxiliary loss to minimize the text-audio difference. For models with augmented training data, we have higher text-audio similarity possibly due to \textit{EN text} and \textit{EN audio} input has the same \textit{ENR text} output during training. Text and audio are represented more similarly in few-shot models than in zero-shot models. This is expected, since few-shot models have seen both \textit{EN audio} and \textit{EN text} being translated to \textit{DE text} during training, while zero-shot models have only seen \textit{EN audio} and \textit{EN text} being translated to different target language during training. Another observation is that the higher the SVCCA score of a zero-shot model, the higher the SVCCA score of the corresponding few-shot model. It suggests that our approaches have already introduced text-audio similarity in the zero-shot settings, but the effect is only clearly shown when we continue to fine-tune with ST data to build few-shot models.
	
	We put the SVCCA scores and the BLEU scores on ST task of few-shot models together to get more insight, as shown in Fig. \ref{fig:models25}b. Observe that most of the scores agree with our hypothesis: more text-audio similarity means higher BLEU score. The only exception is the model with additional opposite data, which has a higher BLEU score but a lower SVCCA score compared to the plain model.
	
	
	We further investigate the exceptional case: additional opposite data, compared to the plain setting. The \textit{EN text} - \textit{DE text} similarity is taken into account. The results are shown in Table \ref{tab:svcca_bi_25}. The SVCCA score of \textit{EN text} - \textit{DE text} for few-shot model in the opposite data setting is 0.607, which is significantly higher than the score of \textit{EN audio} - \textit{EN text} (at 0.267). We also calculate the SVCCA score of \textit{EN text} - \textit{DE text} in the plain setting, and obtain 0.407 points. This indicates that adding opposite data helps the model learn language-independent representation of the data instead of modality-independent representation, and that language-independent representation also helps improving ST performance. A possible reason is that with additional opposite data, we include \textit{EN text $\rightarrow$ DE text} and \textit{DE text $\rightarrow$ EN text} samples during training, hence increase the \textit{EN text} - \textit{DE text} similarity.
	
	\begin{table}[htbp]
		\caption{Modality and language similarity: opposite data versus plain.}
		\begin{center}
			\begin{tabular}{|l|c|c|}
				\hline
				& \begin{tabular}[c]{@{}c@{}}SVCCA \\ EN audio – EN text\end{tabular} & \begin{tabular}[c]{@{}c@{}}SVCCA \\ EN text – DE text\end{tabular} \\ \hline
				Plain few-shot    & \textbf{0.300}                                                     & 0.407                                                             \\ \hline
				Few-shot + opposite data& 0.267                                                              & \textbf{0.607}                                                    \\ \hline
			\end{tabular}
			\label{tab:svcca_bi_25}
		\end{center}
	\end{table}
	
\begin{figure}[htbp]
		\centerline{\includegraphics[width=0.49\textwidth]{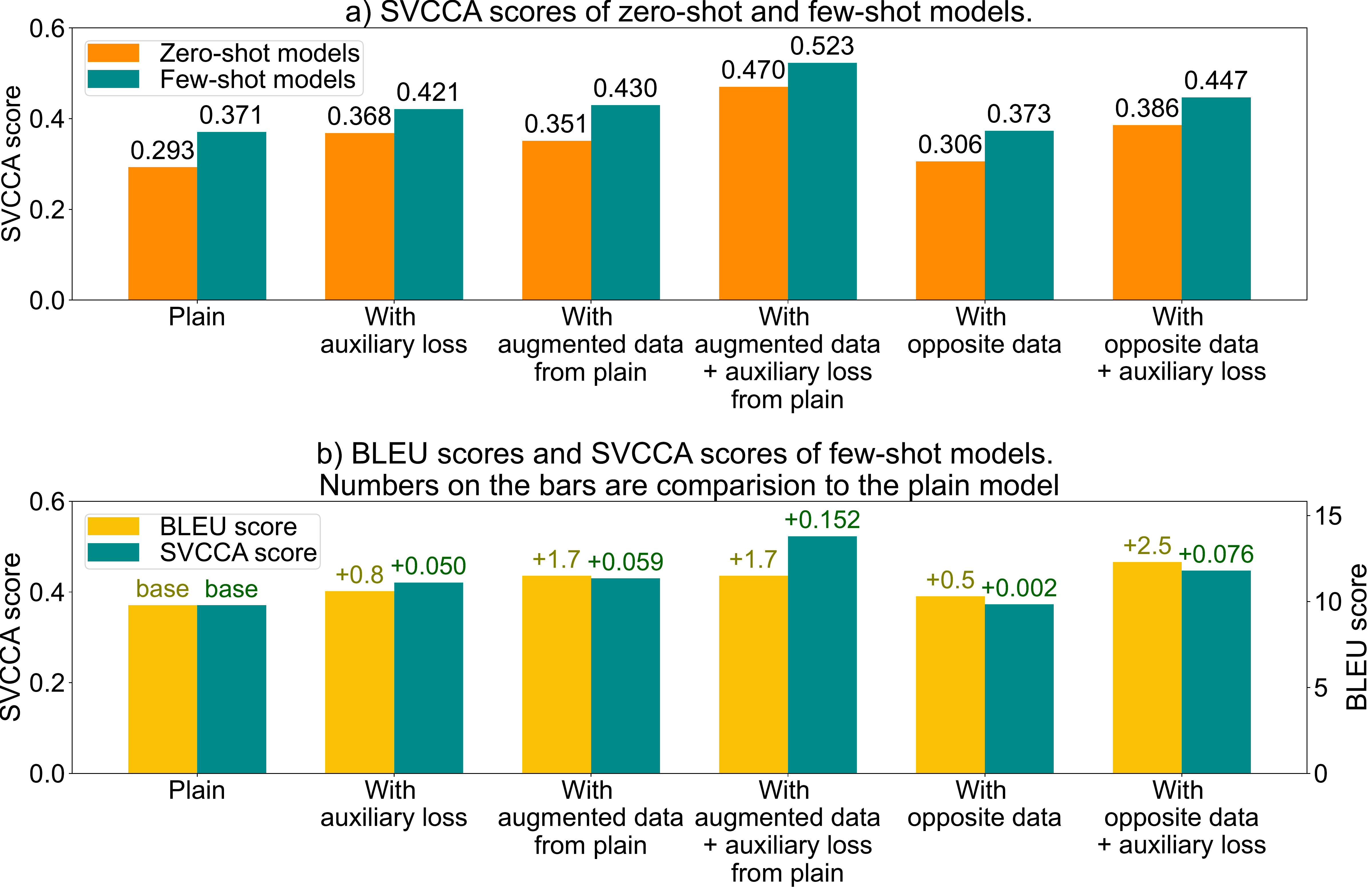}}
		\caption{Analysis of models pre-trained on full data.}
		\label{fig:modelsfull}
	\end{figure}

	The analysis of models trained on full data is shown in Fig. \ref{fig:modelsfull}. Overall, the SVCCA scores of models trained on full data are higher than those trained on 25\% of the data. This is possibly due to the models learn the tasks in a less modality-specific way with a larger amount of training data. We observe similar patterns in the SVCCA scores compared to when the models were trained on 25\% data: most approaches increase the SVCCA score on text-audio similarity; few-shot models have higher SVCCA score than zero-shot models; the higher the SVCCA score of a zero-shot model, the higher the SVCCA score of the corresponding few-shot model; few-shot models with higher SVCCA score have higher BLEU score compared to the plain model. The opposite data approach remains exceptional, where the SVCCA score is approximately the same as the plain setting, but the BLEU score is improved. We also observe that the few-shot model with augmented data and auxiliary loss has the highest SVCCA score on text-audio similarity, yet its ST performance is worse than the few-shot model with opposite data and auxiliary loss. The reason is that models using opposite data can learn language-independent representation of the data, which also helps improving the BLEU score as discussed above. To confirm this, we calculate the SVCCA score of \textit{EN text - DE text} for the two models. The few-shot model using opposite data and auxiliary loss obtains 0.539 points, which is higher than that of the few-shot model using augmented data and auxiliary loss, at 0.397. 
	
	%
	
	The observation in this section also explains why adding auxiliary loss helps improving the performance when combined with additional opposite data, but not for augmented data as shown in Section \ref{sec:combine_exp}. Since the data augmentation approach already helps improving the text-audio similarity, adding auxiliary loss is not significantly useful. In contrast, the opposite data approach does not improve text-audio similarity, hence using auxiliary loss gives a significant improvement.
	
	In addition to the SVCCA score, which measures text-audio similarity on a sentence level through their mean-pooled encoder output, we use another approach to measure text-audio similarity on a token level. We train a linear projection of the encoder output tokens to the modality labels (i.e., text and audio) \cite{DEPI}. Since the number of audio tokens is significantly higher than the number of text tokens, we consider the True Positive Rate (TPR, proportion of audio tokens identified correctly) and the True Negative Rate (TNR, proportion of text tokens identified correctly) of the classification output. Lower rates mean worse classification, meaning text and audio encoder output tokens are more similar. We observe that for all models using auxiliary loss, the TPR is over 99.9\% and the TNR is under 10\%. This means that the classifier has a poor performance where it predicts most of the tokens as audio, which indicates high similarity between text and audio encoder output tokens. On the other hand, for all the models that do not use auxiliary loss, both the TPR and TNR are over 99.9\%, meaning that text and audio encoder output tokens are very distinguishable. Thus, we conclude that using auxiliary loss indeed increases text-audio similarity on a token level.
	
	\section{Conclusions} \label{sec:conclusions}
	The answers to the research questions in Section \ref{sec:intro} are as follows. (1) Our experiment results confirm that cascaded ST is more data-efficient than direct end-to-end ST, since the two cascaded components (i.e., ASR and MT) are more data-efficient. There we emphasize the motivation for zero-shot ST: training an end-to-end model on two data-efficient tasks to perform ST task during inference. (2) We discover that not all approaches on zero-shot multilingual MT can be applied to zero-shot ST. An example is the Disentangling Positional Information approach, which did not work due to the model depth and the text-audio difference. (3) To model the different modalities, we use additional training data and auxiliary loss function, which, by in-depth analysis, prove to enhanced the text-audio similarity in both token and sentence levels.
	
	We find that our approaches are promising for zero-shot ST. We observe that both language-independent and modality-independent representation of the data  improve the ST performance. Our best zero-shot model with opposite training data and auxiliary loss obtains 1.51 BLEU points using no ST data. However, improving zero-shot ST to a practical point remains a difficult task. When a limited amount of ST data is available, our approaches show significant improvements compared to some baselines. We observe improvements of up to +11.8 BLEU points compared to the direct end-to-end ST approach and +3.9 BLEU points compared to models fine-tuned from the ASR task using the same amount of ST data.
	
	For future work, we recommend continuing to enhance models'  modality-independent representation of the data. Our approaches, although prove to have improved text-audio similarity, did not consider the difference in the number of time steps between audio and text. Therefore, we suggest investigating into making the time steps of text and audio more similar or the same. One direction is to use a fixed-size encoder as suggested in \cite{ZS-Quan}. Since the number of time steps of audio input is a lot higher than text input, other directions could be further downsampling audio input, or use Connectionist Temporal Classification (CTC) to compress audio input \cite{gaido2021ctc}. Another approach to encourage text-audio similarity is to represent text input as spoken form using phoneme sequence \cite{tang2021general} and use phone feature for audio input \cite{salesky2020phone}.
	
	\section*{Acknowledgment}
	I am grateful to Dr. Jan Niehues for the guidance throughout the thesis. I also thank Danni Liu and my colleagues at Mediaan for their supports.
	
	\bibliographystyle{IEEEtranN}
	\bibliography{references}
	
	\pagebreak
	\section*{Appendix}
	\subsection{Combination of data augmentation and auxiliary loss}
	\begin{table}[htbp]
		\caption{Zero-shot models using data augmentation in combination with auxiliary loss.}
		\begin{center}
			\begin{tabular}{|r|l|r|r|r|}
				\hline
				\multicolumn{1}{|c|}{\begin{tabular}[c]{@{}c@{}}Data portion\end{tabular}} &
				Model type &
				\multicolumn{1}{l|}{ASR} &
				\multicolumn{1}{l|}{MT} &
				\multicolumn{1}{l|}{Zero-shot ST} \\ \hline
				100\% & Plain zero-shot (ZS)                                                                                            & 28.4 & 32.8 & 0.63 \\ \hline
				100\% & \begin{tabular}[c]{@{}l@{}}ZS + augmented data\\ (fine-tuned from plain)\end{tabular}                    & 27.6 & 32.2 & 0.68 \\ \hline
				100\% & \begin{tabular}[c]{@{}l@{}}ZS + augmented data\\ $\>$ $\>$ $\-$ + auxiliary loss\\ (fine-tuned from plain)\end{tabular} & 27.7 & 32.3 & 0.67 \\ \hline
			\end{tabular}
			\label{tab:ad4_aux}
		\end{center}
	\end{table}
	
	\begin{table}[htbp]
		\caption{Few-shot models from setting with data augmentation in combination with auxiliary loss.}
		\begin{center}
			\begin{tabular}{rlll}
				\hline
				\multicolumn{1}{|c|}{\begin{tabular}[c]{@{}c@{}}Pre-training \\ data portion\end{tabular}} &
				\multicolumn{1}{|c|}{\begin{tabular}[c]{@{}c@{}}Pre-trained \\ model type\end{tabular}} &
				\multicolumn{1}{c|}{\begin{tabular}[c]{@{}c@{}}Fine-tuning data\end{tabular}} &
				\multicolumn{1}{c|}{ST score} \\ \hline
				\multicolumn{1}{|r|}{100\%} &
				\multicolumn{1}{l|}{\begin{tabular}[c]{@{}l@{}}ZS + augmented data \\ (fine-tuned from plain)\end{tabular}} &
				\multicolumn{1}{l|}{10\% ST} &
				\multicolumn{1}{l|}{11.5 \textbf{(+1.7)}} \\ \hline
				\multicolumn{1}{|r|}{100\%} &
				\multicolumn{1}{l|}{\begin{tabular}[c]{@{}l@{}}ZS + augmented data \\ $\>$ $\>$ $\-$ + auxiliary loss\\ (fine-tuned from plain)\end{tabular}} &
				\multicolumn{1}{l|}{10\% ST} &
				\multicolumn{1}{l|}{11.5 \textbf{(+1.7)}} \\ \hline
				\multicolumn{4}{l}{\begin{tabular}[c]{@{}l@{}}Numbers in the brackets are comparison to the corresponding plain setting.\end{tabular}}
			\end{tabular}
			\label{tab:ad4_aux_ft}
		\end{center}
	\end{table}
	
	\subsection{Few-shot models overall performance}
	\begin{table}[htbp]
		\caption{Result on fine-tuning zero-shot models using ST data.}
		\begin{center}
			\begin{tabular}{|r|l|l|l|} 
				\hline
				\multicolumn{1}{|c|}{\begin{tabular}[c]{@{}c@{}}Pre-training\\ data portion\end{tabular}} & \multicolumn{1}{c|}{\begin{tabular}[c]{@{}c@{}}Pre-trained\\ model type\end{tabular}}                                            & \multicolumn{1}{c|}{\begin{tabular}[c]{@{}c@{}}Fine-tuning\\ data\end{tabular}} & \multicolumn{1}{c|}{ST score}                                            \\ 
				\hline
				\multirow{2}{*}{100\%}                                                                & \multirow{2}{*}{Plain zero-shot (ZS)}                                                                                         & 10\% ST                                                                         & \text{ }8.4                                                                      \\
				&                                                                                                                               & 25\% ST                                                                         & 12.4                                                                     \\ 
				\hline
				\multirow{2}{*}{100\%}                                                                & \multirow{2}{*}{ZS + auxiliary loss}                                                                                          & 10\% ST                                                                         & 10.6 \textbf{(+0.8)}                                                     \\
				&                                                                                                                               & 25\% ST                                                                         & 13.2 \textbf{(+0.8)}                                                     \\ 
				\hline
				\multirow{2}{*}{100\%}                                                                & \multirow{2}{*}{\begin{tabular}[c]{@{}l@{}}ZS + augmented data \\ (fine-tuned from plain)\end{tabular}}                       & 10\% ST                                                                         & 11.5 \textbf{(+1.7)}                                                     \\
				&                                                                                                                               & 25\% ST                                                                         & 13.5 \textbf{(+1.1)}                                                     \\ 
				\hline
				\multirow{2}{*}{100\%}                                                                & \multirow{2}{*}{ZS + opposite data}                                                                                           & 10\% ST                                                                         & 10.3 \textbf{(+0.5)}                                                     \\
				&                                                                                                                               & 25\% ST                                                                         & 13.2 \textbf{(+0.8)}                                                     \\ 
				\hline
				\multirow{2}{*}{100\%}                                                                & \multirow{2}{*}{\begin{tabular}[c]{@{}l@{}}ZS + augmented data \\ $\>$ $\>$ $\-$ + auxiliary loss\\ (fine-tuned from plain) \\\end{tabular}} & \begin{tabular}[c]{@{}l@{}}10\% ST \\\end{tabular}                              & \begin{tabular}[c]{@{}l@{}}11.5 \textbf{(+1.7)}\\\end{tabular}  \\
				&                                                                                                                               & \begin{tabular}[c]{@{}l@{}}25\% ST \\$\-$\end{tabular}                              & \begin{tabular}[c]{@{}l@{}}13.7 \textbf{(+1.3)}\\\textbf{}\end{tabular}  \\ 
				\hline
				\multirow{2}{*}{100\%}                                                                & \multirow{2}{*}{\begin{tabular}[c]{@{}l@{}}ZS + opposite data \\ $\>$ $\>$ $\-$ + auxiliary loss\end{tabular}}                               & 10\% ST                                                                         & 12.3 \textbf{(+2.5)}                                                     \\
				&                                                                                                                               & 25\% ST                                                                         & 14.2 \textbf{(+1.8)}                                                     \\ 
				\hline
				\multicolumn{4}{l}{\begin{tabular}[c]{@{}l@{}}Numbers in the brackets are comparison to the plain setting using \\ the same amount of ST data.\end{tabular}}                                                                                                                                                                                               
			\end{tabular}
			\label{tab:finetunefull}
		\end{center}
	\end{table}
	
	\pagebreak
	
	\subsection{Few-shot models fine-tuned on different data}
	\begin{table}[htbp]
		\caption{Few-shot models using ST data versus using ASR, MT, ST data.}
		\begin{center}
			\begin{tabular}{|r|l|l|r|r|r|}
				\hline
				\multicolumn{1}{|c|}{\begin{tabular}[c]{@{}c@{}}Pre-training \\ data portion\end{tabular}} &
				\multicolumn{1}{|c|}{\begin{tabular}[c]{@{}c@{}}Pre-trained \\ model type\end{tabular}} &
				\multicolumn{1}{c|}{\begin{tabular}[c]{@{}c@{}}Fine-tuning\\ data\end{tabular}} &
				\multicolumn{1}{l|}{ASR} &
				\multicolumn{1}{l|}{MT} &
				\multicolumn{1}{c|}{ST} \\ \hline
				100\% & Plain zero-shot & -                                                                         & \textbf{28.4} & \textbf{32.8} & 0.6          \\ \hline
				100\% & Plain zero-shot & 10\% ST                                                                   & 98.2          & 28.3          & \textbf{9.8} \\ \hline
				100\% & Plain zero-shot & \begin{tabular}[c]{@{}l@{}}10\% ASR + \\ 10\% MT + \\ 10\% ST\end{tabular} & 29.5          & 31.4          & 9.1          \\ \hline
			\end{tabular}
			\label{tab:finetuneSTvsALL}
		\end{center}
	\end{table}
\end{document}